\documentclass{ProcISPRAS}
\usepackage[papersize={14.86cm,21cm},left=1.5cm,right=1cm,top=0.8cm,bottom=1cm,includehead,headheight=8pt,heightrounded,headsep=6pt,includefoot,footskip=16pt]{geometry}
\usepackage{amsmath,amssymb}
\usepackage{booktabs,array,graphicx,adjustbox}
\usepackage{tikz}
\usetikzlibrary{arrows.meta,positioning,fit}
\definecolor{prismblue}{HTML}{173C65}
\definecolor{prismteal}{HTML}{087E8B}
\definecolor{prismgold}{HTML}{B86E14}
\definecolor{prismlight}{RGB}{239,244,248}
\definecolor{prismgray}{RGB}{92,101,110}
\tikzset{
  prismbox/.style={draw=prismblue,fill=prismlight,rounded corners,
    align=center,inner sep=5pt,font=\normalsize},
  prismdata/.style={prismbox,draw=prismteal},
  prismresult/.style={prismbox,draw=prismgold},
  prismarrow/.style={-{Latex[length=2mm]},thick,draw=prismgray}
}

\newcommand{\model}{RuModernBERT-ruLaw}
\newcommand{\base}{RuModernBERT-base}
\newcommand{\subhead}[1]{\subsection{#1}}
\definecolor{visBlue}{HTML}{173C65}
\definecolor{visTeal}{HTML}{087E8B}
\definecolor{visAmber}{HTML}{B86E14}
\makeatletter
\newenvironment{fixedtable}{\par\medskip\noindent\begin{minipage}{\linewidth}\def\@captype{table}}{\end{minipage}\par\medskip}
\newenvironment{fixedfigure}{\par\medskip\noindent\begin{minipage}{\linewidth}\def\@captype{figure}\centering}{\end{minipage}\par\medskip}
\makeatother
\DeclareBibliographyDriver{misc}{\printfield{note}\finentry}
\renewcommand{\captionisp}[2]{\caption{#2}}

\title{Domain adaptation of Russian ModernBERT for long legal documents}{Domain adaptation of Russian ModernBERT for long legal documents}
\abstract{We investigate whether continued pretraining on Russian legislative documents improves a Russian ModernBERT encoder on legal text. The adapted model, RuModernBERT-ruLaw, was trained on a corpus reported to contain 304,382 legislative documents and 194,425,905 corpus tokens. Corpus token counts are distinguished from positions produced by the model tokenizer. We compare the original and adapted encoders on a fixed external collection of 1,031 court-decision segments. Both models receive the same hidden positions in each of five masking realizations. At maximum input lengths of 512, 2,048, and 8,192 tokens, mean masked-token cross-entropy decreases by 0.10942, 0.07052, and 0.06604 natural-log units, respectively. The reported 95\% intervals summarize sensitivity to masking on this fixed collection; they do not quantify uncertainty across document collections. A second evaluation addresses legal-entity extraction. The original and adapted models achieve entity-level $F_1$ scores of 0.99852 and 0.99820. However, 99.95\% of test spans have the same normalized surface form and class in the training split. This evaluation therefore provides limited evidence about transfer to previously unseen forms. The paper explains the masking objective, overlapping windows, averaging rules, and exact entity-boundary scoring using editable diagrams and clearly marked illustrative examples. The comparison supports lower masked-token prediction loss for the studied pair of models and collection. It does not isolate the contribution of distant context or establish practical legal utility.}{We investigate whether continued pretraining on Russian legislative documents improves a Russian ModernBERT encoder on legal text. The adapted model, RuModernBERT-ruLaw, was trained on a corpus reported to contain 304,382 legislative documents and 194,425,905 corpus tokens. Corpus token counts are distinguished from positions produced by the model tokenizer. We compare the original and adapted encoders on a fixed external collection of 1,031 court-decision segments. Both models receive the same hidden positions in each of five masking realizations. At maximum input lengths of 512, 2,048, and 8,192 tokens, mean masked-token cross-entropy decreases by 0.10942, 0.07052, and 0.06604 natural-log units, respectively. The reported 95\% intervals summarize sensitivity to masking on this fixed collection; they do not quantify uncertainty across document collections. A second evaluation addresses legal-entity extraction. The original and adapted models achieve entity-level $F_1$ scores of 0.99852 and 0.99820. However, 99.95\% of test spans have the same normalized surface form and class in the training split. This evaluation therefore provides limited evidence about transfer to previously unseen forms. The paper explains the masking objective, overlapping windows, averaging rules, and exact entity-boundary scoring using editable diagrams and clearly marked illustrative examples. The comparison supports lower masked-token prediction loss for the studied pair of models and collection. It does not isolate the contribution of distant context or establish practical legal utility.}
{natural language processing; legal text; ModernBERT; domain adaptation; masked language modeling; long context}
\renewcommand{\authorsep}{, }
\newauthor\authorname{I. Litvak}{I. Litvak}\inst{1}
\newauthor\authorname{D. Gvozdetsky}{D. Gvozdetsky}\inst{2}
\newauthor\authorname{F. Lashkin}{F. Lashkin}\inst{1}
\newauthor\authorname{V. Kirova}{V. Kirova}\inst{2}
\newauthor\authorname{S. Lagutin}{S. Lagutin}\inst{1}
\newauthor\authorname{V. Volf}{V. Volf}\inst{1}
\newauthor\authorname{T. Maksiyan}{T. Maksiyan}\inst{1}
\newauthor\authorname{A. Kostin}{A. Kostin}\inst{3}
\newauthor\authorname{R. Leva}{R. Leva}\inst{1}
\newauthor\authorname{I. Kiselev}{I. Kiselev}\inst{4}
\affil[1]{Independent Researcher}{Independent Researcher}
\affil[2]{National Research Nuclear University MEPhI}{National Research Nuclear University MEPhI}
\affil[3]{Moscow Center for Advanced Studies}{Moscow Center for Advanced Studies}
\affil[4]{University of Waterloo}{University of Waterloo}
\volhead{}\issuehead{}\pageshead{}\yearhead{}
\doi{}
\renewcommand{\tociteen}{}\renewcommand{\tociteru}{}
\begin{document}
\selectlanguage{english}
\maketitleen
\section{Introduction}
In a long legal document, an important condition and a reference to it may be far apart. A model needs to account for words and relationships between document sections to locate relevant passages. Legal language also contains specialized terms and fixed expressions. We continue pretraining a general Russian-language model on legislative texts and evaluate changes in two tasks: predicting hidden tokens and extracting legal spans.

A bidirectional encoder transforms input text into contextual vector representations of its positions using both left and right context. Text is first split into tokens; a rare or complex word may consist of several tokens representing parts of the word. BERT-family encoders remain a practical basis for classification, retrieval, and entity extraction \cite{devlin2019bert}. In masked language modeling, some tokens are hidden. The loss for one hidden token is $-\ln p$, where $p$ is the probability assigned by the model to its correct original value.

General pretraining does not guarantee accurate modeling of language in a particular domain. Continued pretraining on an unlabeled domain corpus, commonly called domain-adaptive pretraining, has systematically improved representations in several domains \cite{gururangan2020dapt}. In the legal domain, LEGAL-BERT compared a general model, continued pretraining, and training from scratch across legal NLP tasks \cite{chalkidis2020legalbert}. Most early legal encoders, however, support only short inputs.

This study investigates domain adaptation of \base{} to Russian legislation while retaining the standard ModernBERT maximum input length of 8192 tokens. The main evaluation uses a court-decision collection that differs from the training corpus in source and document type. Possible content overlap or paraphrased overlap was not specifically excluded. We therefore describe this evaluation as external, rather than independent.

The study addresses three bounded questions. Does continued pretraining reduce cross-entropy at randomly masked positions in external court-decision fragments? Does the direction of the effect persist at maximum input lengths of 512, 2048, and 8192 tokens? Does the available legal-span extraction task distinguish the original and adapted encoders after exact duplicate texts are removed?

The main result is lower masked-token cross-entropy for the adapted model at every tested length and masking realization. Comparing lengths does not isolate a causal effect of distant context because the prefix composition and evaluated tokens also change. Both models achieved $F_1>0.998$ in span extraction, but nearly complete coverage of test forms by training forms limits the discriminative value of this experiment. The study evaluates model losses and one entity-extraction test. It does not assess legal expertise or practical labor savings.

Figure~\ref{fig:research-overview} separates adaptation from the two evaluation branches. External MLM evaluation tests changes in hidden-token probabilities, whereas the span experiment evaluates a separate task after fine-tuning. The result of one branch cannot substitute for that of the other.

\begin{fixedfigure}
\begin{adjustbox}{max width=\linewidth}
\begin{tikzpicture}[font=\normalsize,box/.style={rounded corners,draw=visBlue,fill=visBlue!5,align=center,text width=3.8cm,minimum height=1.7cm,inner sep=5pt},arr/.style={-{Latex[length=2mm]},thick,draw=visBlue}]
\node[box] (base) at (0,0) {Baseline model\\RuModernBERT-base};
\node[box,draw=visTeal,fill=visTeal!7] (train) at (5,0) {Continued pretraining\\on RusLawOD\\304\,382 legislative documents};
\node[box,draw=visAmber,fill=visAmber!6] (law) at (10,0) {RuModernBERT-ruLaw\\input up to 8192 tokens};
\node[box,text width=5.8cm,minimum height=.8cm] (cmp) at (5,-2.05) {Compare baseline\\and adapted models};
\node[box,draw=visTeal,fill=visTeal!7,text width=5.3cm,minimum height=1.6cm] (mlm) at (1.5,-3.8) {MLM: 1031 court fragments\\identical hidden positions\\compare losses};
\node[box,draw=visAmber,fill=visAmber!6,text width=5.3cm,minimum height=1.6cm] (ner) at (8.5,-3.8) {NER: span fine-tuning\\compare boundaries and class\\$P$, $R$, $F_1$};
\draw[arr] (base.east)--(train.west);\draw[arr] (train.east)--(law.west);
\draw[arr] (base.south)--(cmp.north west);\draw[arr] (law.south)--(cmp.north east);
\draw[arr] (cmp.south west)--(mlm.north);\draw[arr] (cmp.south east)--(ner.north);
\end{tikzpicture}
\end{adjustbox}
\captionisp{Domain adaptation and two distinct evaluations. Arrows show experimental stages; masked-token prediction quality does not establish practical legal utility.}{Domain adaptation and two distinct evaluations. Arrows show experimental stages; masked-token prediction quality does not establish practical legal utility.}
\label{fig:research-overview}
\end{fixedfigure}

\subhead{Interpreting model evaluations}
The training task resembles filling gaps in a sentence: the model estimates the probabilities of the original hidden tokens from surrounding text. These probabilities are converted into a numerical loss. A higher probability for the correct token gives a lower loss. Entity extraction poses a different task: locating the boundaries of a relevant text passage and assigning its class. These evaluations assess different model capabilities.

\begin{fixedtable}
\centering\small
\captionisp{Terms used in the diagrams and equations.}{Terms used in the diagrams and equations.}
\begin{tabular}{@{}p{.22\linewidth}p{.70\linewidth}@{}}
\toprule
Term & Meaning in this study\tabularnewline
\midrule
Token & An input-sequence element, often a word or part of a word. One token does not always correspond to one word.\tabularnewline
Context & The tokens provided to the model together with the evaluated position.\tabularnewline
Mask & A selection of hidden positions; this selection changes between repetitions.\tabularnewline
Loss & A value computed from the probability of the correct token or label. At a single position, a lower loss means a higher probability for the correct answer.\tabularnewline
Span & A continuous text segment assigned to a category. We evaluate token boundaries and class within a window.\tabularnewline
Training / test & The training and test splits. Performance on familiar forms differs from transfer to new forms.\tabularnewline
\bottomrule
\end{tabular}
\end{fixedtable}

\section{Related work}

\subhead{Domain adaptation of legal models}
Continued pretraining separates learning general language patterns from adaptation to the distribution of a target corpus \cite{gururangan2020dapt}. LEGAL-BERT showed that the choice between using a general model, continuing pretraining, and training from scratch requires experimental evaluation: recommendations from a general domain do not always transfer to legal tasks \cite{chalkidis2020legalbert}. The later MultiLegalPile corpus combined material in 24 languages from 17 jurisdictions and was used to train RoBERTa and Longformer models, emphasizing both language and jurisdictional specificity \cite{niklaus2024multilegalpile}.

The Russian Legislative Corpus (RusLawOD) is an important source for Russian. Version~3 of arXiv:2406.04855, dated 27 April 2026, describes a 1991--2025 snapshot containing 304\,382 documents and 194\,425\,905 tokens \cite{saveliev2024corpus}. The corpus is available through Hugging Face \cite{ruslawodcard}. These statistics refer to the snapshot described in version~3.

\subhead{Long-context encoders}
Full self-attention has quadratic cost in sequence length. Longformer combines local sliding-window attention with designated global attention, yielding linear scaling for most operations \cite{beltagy2020longformer}. Based on this architecture, Lawformer was further pretrained on Chinese court documents of up to 4096 tokens and improved performance on tasks involving long legal texts \cite{xiao2021lawformer}.

ModernBERT develops the bidirectional encoder with a standard context length of 8192 tokens, alternating global and local attention, rotary positional representations, and modern computational optimizations \cite{warner2024modernbert}. The model studied here adapts ModernBERT to Russian legislation and separately checks whether the adaptation effect persists at three input lengths.

\section{Data and method}

\subhead{Corpus and training objective}
The initial model was \texttt{deepvk/RuModernBERT-base}. Training used the \texttt{textIPS} field of RusLawOD. Version~3 of the corpus publication, dated 27 April 2026, reports $D=304\,382$ documents and $N=194\,425\,905$ tokens \cite{saveliev2024corpus}. The corpus is available through Hugging Face \cite{ruslawodcard}. Here $N$ counts corpus tokens. The number of model-token positions depends on the model tokenizer, windowing, and overlap, and is a different quantity. Documents were split into windows of up to $L_{\max}=8192$ tokens with an overlap of 512 tokens.

Let a minibatch contain $B$ tokenized windows $W_b=(x_{b,1},\ldots,x_{b,L_b})$. A set of masked positions $M_b$, which may be empty, is selected for each window. The input after masking and token replacement is denoted by $\widetilde W_b$. The training loss averages over all masked tokens in the minibatch, provided that $\sum_{b=1}^{B}|M_b|>0$:
\begin{equation}
 \widehat{\mathcal L}_{\mathrm{train}}(\theta)
 =-\frac{1}{\sum_{b=1}^{B}|M_b|}
 \sum_{b=1}^{B}\sum_{i\in M_b}
 \log p_\theta(x_{b,i}\mid \widetilde W_b).
 \label{eq:mlm}
\end{equation}
With dynamic masking, each eligible position is selected with probability $q=0.15$. An empty $M_b$ for one window does not invalidate the formula if the minibatch contains at least one masked position. If the entire minibatch mask is empty, the loss in Eq.~\eqref{eq:mlm} is undefined. Throughout the paper, $\log=\ln$, so losses and cross-entropies are expressed in nats. The external evaluation below uses a different aggregation order: first over masked tokens in a fragment with a batch size of one, then with equal weights over fragments and masking realizations.

Let $L$ denote sequence length, $w$ local-window width, $d$ hidden dimension, $h$ the number of heads, $n_\ell$ and $n_g$ the numbers of local and global layers, $B$ batch size, $P$ the number of model parameters, and $p$ the bytes per stored element. The local and global attention products have cost of order
\begin{equation}
 C_{\mathrm{attn}}=O\!\left(Bd\,[n_\ell Lw+n_gL^2]\right).
 \label{eq:attention}
\end{equation}
The head count $h$ affects the shape and size of intermediate matrices. At a fixed total hidden dimension $d$, the leading arithmetic summed over heads yields the estimate above. Local layers scale linearly and global layers quadratically with $L$. This estimate excludes linear projections, feed-forward blocks, communication, and implementation overhead.

\begin{fixedtable}
\centering\small
\captionisp{Attention cost and scaling variables. Structural operation and memory estimates are distinct from measured runtime and peak memory.}{Attention cost and scaling variables. Structural operation and memory estimates are distinct from measured runtime and peak memory.}
\label{tab:attention-resources}
\begin{tabular}{@{}p{0.20\linewidth}p{0.34\linewidth}p{0.34\linewidth}@{}}
\toprule
Component & Time / arithmetic & Memory\tabularnewline
\midrule
Local attention & $O(Bn_\ell Lwd)$ & $O(pBhn_\ell Lw)$ when matrices for all local layers are explicitly stored\tabularnewline
Global attention & $O(Bn_gL^2d)$ & $O(pBhn_gL^2)$ when matrices for all global layers are explicitly stored\tabularnewline
Parameters and states & Depend on optimizer implementation & $O(pP)$ for one parameter representation; training stores additional states\tabularnewline
FlashAttention & Does not remove quadratic arithmetic in global layers & Attention-kernel working memory $O(pBLd)$ is linear in $L$ at fixed $d,h$ and does not require a full $L\times L$ matrix\tabularnewline
\bottomrule
\end{tabular}
\end{fixedtable}

The factors $n_\ell$ and $n_g$ in the explicit attention-weight memory estimates sum the stored layer matrices, as in training with retained activations. During sequential model execution, peak memory may contain only the current layer matrix and omit the layer-count factor. The FlashAttention estimate concerns attention working memory. Total model activations, parameters, gradients, optimizer states, padding, buffers, and inter-accelerator communication are accounted for separately and are not bounded by that estimate. The experimental materials provide no measured training duration, throughput, or peak memory. The table therefore describes asymptotic scaling rather than a measured advantage of this run.

Figure~\ref{fig:document-windows} shows a windowed representation of a long document. Overlap retains some local context near boundaries but includes the same positions in neighboring windows. Corpus token counts should therefore not be equated with the number of positions actually processed after windowing.

\begin{fixedfigure}
\begin{adjustbox}{max width=\linewidth}
\begin{tikzpicture}[x=.65cm,y=1cm,font=\normalsize]
\node[anchor=east,text=visBlue] at (-.5,2.2) {document};
\foreach \i in {1,...,14}{
 \draw[black!35,fill=visBlue!5] (\i-1,1.9) rectangle (\i,2.5);
 \node[font=\small] at (\i-.5,2.2) {\i};
}
\node[anchor=east,text=visBlue] at (-.5,1.1) {window 1};
\foreach \i in {1,...,8}{
 \ifnum\i>6 \def\cc{visAmber!20}\else\def\cc{visBlue!14}\fi
 \draw[visBlue,fill=\cc] (\i-1,.8) rectangle (\i,1.4);
 \node at (\i-.5,1.1) {\i};
}
\node[anchor=east,text=visTeal] at (-.5,0) {window 2};
\foreach \i in {7,...,14}{
 \ifnum\i<9 \def\cc{visAmber!20}\else\def\cc{visTeal!15}\fi
 \draw[visTeal,fill=\cc] (\i-1,-.3) rectangle (\i,.3);
 \node at (\i-.5,0) {\i};
}
\draw[-{Latex[length=2mm]},visAmber,thick] (6.5,.72)--(6.5,.38);
\draw[-{Latex[length=2mm]},visAmber,thick] (7.5,.72)--(7.5,.38);
\node[align=center,text width=12cm] at (6,-1.0) {Illustrative diagram: $L=8$, overlap of 2 positions (7 and 8).\\Run settings: MLM $8192/512$; NER $8192/1024$\\(maximum length / overlap).};
\end{tikzpicture}
\end{adjustbox}
\captionisp{Overlapping windows retain neighbouring boundary tokens. The 14 numbered positions are an illustrative example; positions 7 and 8 belong to both windows. The actual maximum-length/overlap settings are 8192/512 for language modelling and 8192/1024 for entity extraction.}{Overlapping windows retain neighbouring boundary tokens. The 14 numbered positions are an illustrative example; positions 7 and 8 belong to both windows. The actual maximum-length/overlap settings are 8192/512 for language modelling and 8192/1024 for entity extraction.}
\label{fig:document-windows}
\end{fixedfigure}

Figure~\ref{fig:masked-token-example} explains one term of the MLM loss: a lower probability assigned to the correct hidden token gives a larger $-\ln p$. The training loss aggregates these terms over masked minibatch positions according to Eq.~\eqref{eq:mlm}.

\begin{fixedfigure}
\begin{adjustbox}{max width=\linewidth}
\begin{tikzpicture}[font=\normalsize,box/.style={rounded corners,draw=visBlue,fill=visBlue!5,align=center,text width=3.7cm,minimum height=1.8cm,inner sep=5pt},arr/.style={-{Latex[length=2mm]},thick,draw=visBlue}]
\node[box] (source) at (0,0) {1. Original phrase\\Court considered\\the motion};
\node[box,draw=visTeal,fill=visTeal!7] (masked) at (4.7,0) {2. Hide a position\\Court considered\\\texttt{[MASK]}};
\node[box,draw=visAmber,fill=visAmber!7] (pred) at (9.4,0) {3. Target probability\\the motion\\$p_\theta$};
\node[box,draw=visAmber,fill=visAmber!7,text width=5cm,minimum height=1cm] (loss) at (9.0,-2) {4. Loss: $-\ln p_\theta$, nat\\higher probability --- lower loss};
\draw[arr] (source.east)--(masked.west);\draw[arr] (masked.east)--(pred.west);\draw[arr] (pred.south)--(loss.north);
\end{tikzpicture}
\end{adjustbox}
\captionisp{An illustrative hidden-word prediction task and its loss. Words and probabilities are examples rather than model measurements; actual tokenization may split a word, and training corruption is not limited to replacement by the mask symbol.}{An illustrative hidden-word prediction task and its loss. Words and probabilities are examples rather than model measurements; actual tokenization may split a word, and training corruption is not limited to replacement by the mask symbol.}
\label{fig:masked-token-example}
\end{fixedfigure}

In an illustrative numerical example, a correct-token probability of $0.5$ gives a loss of $-\ln(0.5)\approx0.693$ nats; a probability of $0.8$ gives $0.223$ nats. These are hypothetical probabilities, not measured model responses. The example illustrates that a higher correct-token probability gives a lower loss.

\subhead{How length affects computation}
Attention connects token positions: information from other positions contributes to the representation of one position. A local layer restricts these connections to neighbors; a global layer connects all pairs. Figure~\ref{fig:attention-grid} represents these patterns as a narrow band and a full square. At fixed local-window width, doubling length approximately doubles the local pair count and quadruples the global pair count. Positions at the boundaries have fewer neighbors, which explains the approximation. The asymptotic estimates are given in Eq.~\eqref{eq:attention} and Table~\ref{tab:attention-resources}.

\begin{fixedfigure}
\centering
\begin{adjustbox}{max width=\linewidth}
\begin{tikzpicture}[font=\normalsize]
\node[font=\normalsize,text=visBlue] at (1.6,4.25) {Local attention};
\node[font=\normalsize,text=visTeal] at (7.0,4.25) {Global attention};
\foreach \i in {0,...,7}{\foreach \j in {0,...,7}{
 \pgfmathtruncatemacro{\near}{abs(\i-\j)<=1}
 \ifnum\near=1 \fill[visBlue!24] (\j*.4,3.6-\i*.4) rectangle ++(.4,-.4);\fi
 \draw[black!25,line width=.2pt] (\j*.4,3.6-\i*.4) rectangle ++(.4,-.4);
 \fill[visTeal!25] (5.4+\j*.4,3.6-\i*.4) rectangle ++(.4,-.4);
 \draw[black!25,line width=.2pt] (5.4+\j*.4,3.6-\i*.4) rectangle ++(.4,-.4);
}}
\foreach \i in {1,...,8}{
 \node[font=\scriptsize] at (.4*\i-.2,3.85) {\i};
 \node[font=\scriptsize] at (-.25,3.8-.4*\i) {\i};
 \node[font=\scriptsize] at (5.4+.4*\i-.2,3.85) {\i};
 \node[font=\scriptsize] at (5.15,3.8-.4*\i) {\i};
}
\node[align=center,text width=4cm] at (1.6,-.2) {Position attends to\\neighboring positions};
\node[align=center,text width=4cm] at (7.0,-.2) {Position attends to\\all positions};
\node[draw=visAmber,fill=visAmber!7,rounded corners,align=center,text width=3.2cm,inner sep=6pt] at (11.3,2.2) {When doubling $L$\\at fixed $w$:\\[3pt]local pairs approximately $\times2$\\[3pt]global pairs $\times4$};
\end{tikzpicture}
\end{adjustbox}
\captionisp{Local and global attention on an illustrative eight-position sequence with local width three; this is not the run configuration. Rows and columns represent token positions, and filled cells denote accessible pairs. Global layers connect distant positions. FlashAttention reduces storage of intermediate matrices while preserving computation of global pairs. With fixed local width, doubling sequence length approximately doubles the local pair count and quadruples the full pair count, ignoring boundary effects; these are not hardware measurements.}{Local and global attention on an illustrative eight-position sequence with local width three; this is not the run configuration. Rows and columns represent token positions, and filled cells denote accessible pairs. Global layers connect distant positions. FlashAttention reduces storage of intermediate matrices while preserving computation of global pairs. With fixed local width, doubling sequence length approximately doubles the local pair count and quadruples the full pair count, ignoring boundary effects; these are not hardware measurements.}
\label{fig:attention-grid}
\end{fixedfigure}

Computational cost concerns the number of operations, whereas memory concerns the amount of data stored simultaneously. The order $O(L^2)$ describes growth of the leading term as $L$ increases; it does not specify elapsed seconds. FlashAttention computes attention in blocks and reduces intermediate storage. It retains the global pairs and does not provide a measured runtime or memory footprint for this run.

\subhead{External MLM evaluation}
The external sample was constructed from an open dataset of anonymized court decisions \cite{sudreshcard}. Filtering and removal of exact duplicates yielded $K=1031$ fragments. Prefixes of length $L\in\{512,2048,8192\}$ were formed from a single tokenization while retaining the final special token. The training corpus RusLawOD and the evaluation collection differ in source and document type, but close paraphrased matches between them were not specifically checked.

For masking seeds $s\in S=\{11,23,42,67,101\}$, both models receive the same position sets $M_{k,s,L}$. Evaluation with equal fragment weights requires $|M_{k,s,L}|>0$ for every evaluated fragment; the evaluation procedure must ensure or check this condition. Let $\widetilde W_{k,s,L}$ denote inputs with identical masking and token replacements for both models, and let
\begin{equation}
 h_{m,k,s,L}=-\frac{1}{|M_{k,s,L}|}\sum_{i\in M_{k,s,L}}
 \log p_m(x_{k,i}\mid\widetilde W_{k,s,L}),\qquad
 \widehat H_{m,s,L}=\frac{1}{K}\sum_{k=1}^{K}h_{m,k,s,L},
 \label{eq:external-estimator}
\end{equation}
where $m\in\{\mathrm{base},\mathrm{law}\}$ and $K=1031$. Each fragment thus has the same weight regardless of its masked-position count. A batch size of one fragment aligns the implemented procedure with this definition.

\begin{fixedfigure}
\centering
\begin{adjustbox}{max width=\linewidth}
\begin{tikzpicture}[font=\normalsize,box/.style={draw=visBlue,fill=visBlue!5,rounded corners,align=center,text width=5.3cm,inner sep=6pt},arr/.style={-{Latex[length=2mm]},thick,draw=visBlue}]
\node[box] (a) at (0,0) {Fragment A: 1 hidden position\\loss $1.0$ nat\\fragment mean $1.0$};
\node[box] (b) at (6.4,0) {Fragment B: 3 hidden positions\\losses $0.2;\;0.2;\;0.2$ nat\\fragment mean $0.2$};
\node[box,fill=visTeal!7,draw=visTeal,text width=5.5cm] (c) at (3.2,-2.2) {External evaluation:\\equal fragment weights\\$(1.0+0.2)/2=0.6$ nat};
\draw[arr] (a.south)--(c.north west); \draw[arr] (b.south)--(c.north east);
\node[align=center,text width=12cm] at (3.2,-3.6) {Averaging over all four hidden positions gives a different value:\\$(1.0+3\cdot0.2)/4=0.4$ nat. Fragment B would receive three times the weight.};
\end{tikzpicture}
\end{adjustbox}
\captionisp{Averaging order changes the interpretation. All values are illustrative. External evaluation first averages losses within each fragment and then gives equal weight to fragments; the training loss in Eq.~\eqref{eq:mlm} weights masked positions.}{Averaging order changes the interpretation. All values are illustrative. External evaluation first averages losses within each fragment and then gives equal weight to fragments; the training loss in Eq.~\eqref{eq:mlm} weights masked positions.}
\label{fig:averaging-example}
\end{fixedfigure}

The paired difference for one masking realization is
\begin{equation}
 d_{s,L}=\widehat H_{\mathrm{base},s,L}-\widehat H_{\mathrm{law},s,L}
 \label{eq:paired}
\end{equation}
and is positive when the adapted model has lower loss. The final means are $\widehat H_{m,L}=n^{-1}\sum_{s\in S}\widehat H_{m,s,L}$ and $\overline d_L=n^{-1}\sum_{s\in S}d_{s,L}$, where $n=|S|=5$. The sample standard deviation of paired differences is
\begin{equation}
 s_{d,L}=\sqrt{\frac{1}{n-1}\sum_{s\in S}(d_{s,L}-\overline d_L)^2}.
 \label{eq:paired-sd}
\end{equation}
A confidence interval was computed from the five aggregated differences:
\begin{equation}
 \overline d_L\pm t_{0.975;n-1}\frac{s_{d,L}}{\sqrt n},\qquad t_{0.975;4}=2.776.
 \label{eq:ci}
\end{equation}
This $t$-interval assumes independent masking realizations and an approximately normal distribution of aggregated differences $d_{s,L}$. With $n=5$, normality is practically impossible to assess from the observations themselves. The interval describes sensitivity of the aggregated result to random masking on this fixed collection. Five masks are not five independent corpora. The interval excludes variability due to document selection and does not estimate transfer to arbitrary legal texts.

\begin{fixedfigure}
\begin{adjustbox}{max width=\linewidth}
\begin{tikzpicture}[font=\normalsize,box/.style={draw=visBlue,fill=visBlue!5,rounded corners,align=center,text width=5.4cm,inner sep=5pt,minimum height=1cm},arr/.style={-{Latex[length=2mm]},thick,draw=visBlue}]
\node[box,draw=visTeal,fill=visTeal!7] (data) at (0,0) {The same 1031 court fragments\\and the same hidden positions\\for both models};
\node[box,text width=4cm] (base) at (-3.7,-1.9) {Before adaptation\\$\widehat H_{\mathrm{base},s,L}$};
\node[box,draw=visAmber,fill=visAmber!7,text width=4cm] (law) at (3.7,-1.9) {After adaptation\\$\widehat H_{\mathrm{law},s,L}$};
\node[box,draw=visAmber,fill=visAmber!7] (diff) at (0,-3.4) {Paired difference\\$d_{s,L}=\widehat H_{\mathrm{base},s,L}-\widehat H_{\mathrm{law},s,L}$};
\node[box,draw=visTeal,fill=visTeal!7,text width=6.8cm,minimum height=1.2cm] (scope) at (0,-5.2) {Repeat with five masks\\Mean of five paired differences\\Sensitivity to masking};
\draw[arr] (data.south west)--(base.north);\draw[arr] (data.south east)--(law.north);
\draw[arr] (base.south)--(diff.north west);\draw[arr] (law.south)--(diff.north east);\draw[arr] (diff.south)--(scope.north);
\end{tikzpicture}
\end{adjustbox}
\captionisp{Paired evaluation uses the same 1031 court fragments and hidden positions for both models at each length and masking seed. Five masks measure masking sensitivity on a fixed collection, not variation across independent corpora.}{Paired evaluation uses the same 1031 court fragments and hidden positions for both models at each length and masking seed. Five masks measure masking sensitivity on a fixed collection, not variation across independent corpora.}
\label{fig:paired-mlm}
\end{fixedfigure}

\noindent\textit{Illustrative example.} Prediction difficulty may differ if one mask hides the word for ``motion'' and another hides the word for ``court''. The original and adapted models are therefore compared using the same mask. Repeating the evaluation with five masks tests sensitivity to hidden-position selection; it does not turn one fragment into five independent documents.

We also report $\exp(\widehat H)$, a monotonic transformation of mean cross-entropy measured in nats using the natural logarithm. It is not the usual autoregressive perplexity of a complete sequence. Since $\widehat H_{m,s,L}$ is first averaged over masks, this quantity also differs from the mean $n^{-1}\sum_s\exp(\widehat H_{m,s,L})$. The relative reduction in the exponential of mean cross-entropy follows directly from the paired difference:
\begin{equation}
 r_L=1-\frac{\exp(\widehat H_{\mathrm{law},L})}{\exp(\widehat H_{\mathrm{base},L})}=1-\exp(-\overline d_L).
 \label{eq:relative}
\end{equation}

\subhead{Legal-span extraction evaluation}
The additional task provides character intervals for five legal categories with numeric identifiers $2,4,9,13,17$ \cite{nercard}. Intervals were converted to BIO labels: \texttt{B} marks the first token of an entity, \texttt{I} its continuation, and \texttt{O} a position outside any entity. The set $\Omega$ contains all token positions eligible for labeling that are neither special nor padding tokens; it therefore includes positions labeled \texttt{O}. Special and padding tokens were assigned the index $-100$ and excluded from the loss:
\begin{equation}
 \mathcal L_{\mathrm{span}}(\phi)=-\frac{1}{|\Omega|}\sum_{t\in\Omega}\log p_\phi(z_t\mid W).
  \label{eq:span}
  \end{equation}
Documents were processed in windows of up to 8192 tokens with an overlap of 1024. The \texttt{seqeval} implementation evaluated each window as a separate sequence. Predictions from overlapping windows were not explicitly merged at the original-document level. Microaveraged precision, recall, and $F_1$ were computed from BIO sequences using exact matches of token boundaries and entity class. Here $TP$ denotes correctly recovered entities, $FP$ predicted entities without an exact reference match, and $FN$ reference entities not recovered exactly:
  \begin{equation}
   P=\frac{TP}{TP+FP},\qquad R=\frac{TP}{TP+FN},\qquad F_1=\frac{2PR}{P+R}.
    \label{eq:f1}
    \end{equation}

Before splitting, rows were grouped by the SHA-256 hash of normalized text. Of 97\,668 rows, 15\,783 duplicate rows and 109 rows belonging to 32 texts with conflicting annotations were removed. The remaining 81\,776 unique texts were deterministically divided into 65\,382 training, 8133 validation, and 8261 test examples. Exact texts do not overlap between splits.

This split prevents overlap of identical normalized texts but does not exclude recurring span surface forms. Window-level token evaluation also differs from recovering exact character intervals in an original document: tokenization, window boundaries, and the absence of explicit merging limit this correspondence.

    \begin{fixedfigure}
\centering
\begin{adjustbox}{max width=\linewidth}
\begin{tikzpicture}[
 node distance=8mm and 12mm,
 arr/.style={prismarrow}]
\node[prismdata] (rows) {97\,668 rows};
\node[prismbox,right=of rows] (dedup) {group by SHA-256\\of normalized text};
\node[prismresult,right=of dedup] (split) {disjoint\\exact training/test texts};
\node[prismresult,below=of split] (forms) {span surface forms\\may recur between splits};
\draw[arr] (rows)--(dedup);
\draw[arr] (dedup)--(split);
\draw[arr] (split)--node[right,font=\footnotesize]{does not prevent}(forms);
\end{tikzpicture}
\end{adjustbox}
\captionisp{Exact-text deduplication does not prevent overlap of normalized entity forms between training and test splits.}{Exact-text deduplication does not prevent overlap of normalized entity forms between training and test splits.}
\label{fig:text-split-span-forms}
\end{fixedfigure}

Figure~\ref{fig:bio-example} illustrates why entity-level evaluation requires matching both span boundaries and class. Moving the beginning of an entity by one token makes the prediction incorrect under exact matching. The values $TP=3$, $FP=1$, and $FN=1$ are used only to illustrate Eq.~\eqref{eq:f1}; they are not experimental results from this study.

\begin{fixedfigure}
\begin{adjustbox}{max width=\linewidth}
\begin{tikzpicture}[font=\normalsize,
 cell/.style={draw=visBlue,fill=visBlue!5,rounded corners,minimum width=1.8cm,minimum height=.55cm,inner sep=3pt,align=center}]
\foreach \y/\lbl in {0/token,-.9/reference,-1.8/error}{\node[anchor=east,text=visBlue] at (-1.3,\y) {\lbl};}
\foreach \x/\word/\gold/\pred in {0/Court/O/O,2.2/of city/B-X/O,4.4/N/I-X/B-X,6.6/ruled/O/O}{
 \node[cell] at (\x,0) {\word};
 \node[cell,draw=visTeal,fill=visTeal!8] at (\x,-.9) {\gold};
 \node[cell,draw=visAmber,fill=visAmber!8] at (\x,-1.8) {\pred};
}
\node[draw=visBlue,fill=visBlue!4,rounded corners,align=center,text width=12cm,inner sep=6pt] (exact) at (2.8,-3.15) {Reference: ``of city N. Prediction: only ``N.\\Boundaries differ: one $FP$ and one $FN$, no $TP$.};
\node[draw=visAmber,fill=visAmber!7,rounded corners,align=center,text width=12cm,inner sep=6pt] at (2.8,-4.6) {Separate illustrative aggregate: $TP=3$, $FP=1$, $FN=1$.\\$P=3/(3+1)=0.75$, $R=3/(3+1)=0.75$, $F_1=0.75$.};
\end{tikzpicture}
\end{adjustbox}
\captionisp{Illustrative entity boundaries in BIO notation: O is outside, B-X starts an entity, and I-X continues it. X is a fictional category and tokenization is illustrative. A boundary error and the separate aggregate calculation are teaching examples, not experimental results.}{Illustrative entity boundaries in BIO notation: O is outside, B-X starts an entity, and I-X continues it. X is a fictional category and tokenization is illustrative. A boundary error and the separate aggregate calculation are teaching examples, not experimental results.}
\label{fig:bio-example}
\end{fixedfigure}

These example counts give $P=R=F_1=3/4=0.75$. Overall token accuracy does not describe this boundary error: most words could be labeled correctly. The counts $TP$, $FP$, and $FN$ clarify what is assessed, whereas $P$, $R$, and $F_1$ provide normalized summaries for comparison.

\section{Computational experiment}
Training was performed on one node of the National Research Nuclear University MEPhI high-performance computing cluster with two AMD EPYC 7542 processors, 512~GiB of RAM, and eight NVIDIA A100 PCIe GPUs with 40~GB of memory each. The software versions were Python 3.11.15, PyTorch 2.6.0 with CUDA 12.4, Transformers 4.57.3, and DeepSpeed 0.19.1.

Continued pretraining used BF16, FlashAttention-2 \cite{dao2023flashattention2}, and distributed optimization with DeepSpeed ZeRO-2 \cite{rajbhandari2020zero}. The actual run used four sequences per accelerator and one gradient-accumulation step, giving
\begin{equation}
 B_{\mathrm{global}}=N_{\mathrm{GPU}}B_{\mathrm{device}}A=8\cdot4\cdot1=32.
 \label{eq:batch}
\end{equation}
The upper bound on token positions per optimizer step is $32\cdot8192=262\,144$. The actual number of useful positions depends on sequence lengths, padding, and how the implementation handles it. AdamW \cite{loshchilov2019adamw} was used with an initial learning rate of $7\cdot10^{-5}$. Of the planned 20 epochs, early stopping occurred at approximately epoch 16.38. The minimum validation loss of 0.133617 was recorded at step 200\,000. The run identifier was \texttt{mlm-20e-bs4-lr7e5}. The public repository, branch \texttt{master}, commit \texttt{b3b3cba1e03018a36e1c2e478f34fa3ca2bd0806}, contains the code, \texttt{logs.gz}, and \texttt{evidence-index.json} inspected during preparation of the full source version \cite{repository}. At that inspection, the manifest-linked file \path{artifacts/reproducibility/metrics/outputs/evidence/mlm-context/summary.json} returned HTTP 404. Public code and partial logs therefore do not establish completeness of the raw metrics or independent numerical reproduction. The published model package is available on Hugging Face \cite{modelcard}; its model-card benchmark table describes other internal runs and should not be substituted for the paired evaluation and cleaned NER comparison reported here.

For span extraction, the original and adapted encoders were each fine-tuned for three epochs on the same split with identical hyperparameters: a global batch size of 8 and an initial learning rate of $3\cdot10^{-5}$.

\section{Results and discussion}

\subhead{MLM evaluation results}
\begin{fixedtable}
\centering\small
\captionisp{External masked-token evaluation at different maximum lengths. Losses are averaged over five masks and expressed in natural-log units.}{External masked-token evaluation at different maximum lengths. Losses are averaged over five masks and expressed in natural-log units.}
\label{tab:mlm-context}
\begin{tabular}{@{}rccccc@{}}
\toprule
$L$ & $\widehat H_{\mathrm{base},L}$ & $\widehat H_{\mathrm{law},L}$ & $\overline d_L$ & 95\% CI & $100r_L$, \%\\
\midrule
512  & 0.77487 & 0.66545 & 0.10942 & [0.10281; 0.11603] & 10.36\\
2048 & 0.54954 & 0.47903 & 0.07052 & [0.06848; 0.07255] & 6.81\\
8192 & 0.51502 & 0.44898 & 0.06604 & [0.06468; 0.06740] & 6.39\\
\bottomrule
\end{tabular}
\end{fixedtable}

\begin{fixedfigure}
\centering
\begin{adjustbox}{max width=\linewidth}
\begin{tikzpicture}[font=\normalsize,x=80cm,y=1cm]
\draw[black!45] (0,0)--(.125,0);
\foreach \x/\s in {0/0, .02/{0.02},.04/{0.04},.06/{0.06},.08/{0.08},.10/{0.10},.12/{0.12}}{
 \draw[black!30] (\x,0)--(\x,3.2); \node[below] at (\x,0) {$\s$};}
\node[below,align=center] at (.063,-.6) {Paired loss difference, nats: $\overline d_L=\widehat H_{\mathrm{base},L}-\widehat H_{\mathrm{law},L}$};
\foreach \y/\len/\mid/\lo/\hi/\lab in {3/512/.10942/.10281/.11603/{0.10942},2/2048/.07052/.06848/.07255/{0.07052},1/8192/.06604/.06468/.06740/{0.06604}}{
 \node[left,text=visBlue] at (-.004,\y) {$L=\len$};
 \draw[visTeal,very thick] (\lo,\y)--(\hi,\y);
 \draw[visTeal,thick] (\lo,\y-.10)--(\lo,\y+.10);
 \draw[visTeal,thick] (\hi,\y-.10)--(\hi,\y+.10);
 \fill[visBlue] (\mid,\y) circle[radius=2pt];
 \node[above=4pt,text=visBlue] at (\mid,\y) {$\lab$};
}
\node[align=left,text width=11cm,anchor=west] at (-.008,4.25) {\textcolor{visBlue}{All three mean differences are positive:} the adapted model has lower loss on this collection.};
\end{tikzpicture}
\end{adjustbox}
\captionisp{Original language-modelling results without new measurements. Points show mean paired loss differences; segments show separate 95\% $t$-intervals across five masks of the fixed collection of 1031 fragments. Zero denotes equal mean losses. These intervals describe variation due to masking. The comparison does not isolate distant-context effects.}{Original language-modelling results without new measurements. Points show mean paired loss differences; segments show separate 95\% $t$-intervals across five masks of the fixed collection of 1031 fragments. Zero denotes equal mean losses. These intervals describe variation due to masking. The comparison does not isolate distant-context effects.}
\label{fig:mlm-results-ci}
\end{fixedfigure}

A separate pointwise 95\% $t$-interval across five masks is reported for each length $L$. The three intervals were not adjusted for simultaneous coverage across all three lengths.

The adapted model had lower loss at every length and in each of the five masking realizations. At a maximum input length of 8192 tokens, cross-entropy decreased by 0.06604 on average, corresponding to a 6.39\% reduction in its exponential. This percentage describes the change in $\exp(\widehat H)$, rather than in cross-entropy itself or the fraction of correct predictions. The effect is larger at $L=512$, but three points do not establish a monotonic relationship between gain and length: both the available context and the evaluated prefix tokens change.

\subhead{Span extraction results}
\begin{fixedtable}
\centering
\captionisp{Legal-entity extraction on the cleaned test split.}{Legal-entity extraction on the cleaned test split.}
\label{tab:ner}
\begin{tabular}{@{}lccc@{}}
\toprule
Encoder & Precision & Recall & $F_1$\\
\midrule
\base{}  & 0.99841 & 0.99863 & 0.99852\\
\model{} & 0.99794 & 0.99846 & 0.99820\\
\bottomrule
\end{tabular}
\end{fixedtable}

The original and adapted encoders achieved $F_1=0.99852$ and $F_1=0.99820$, respectively; their test losses were $6.58\cdot10^{-5}$ and $6.57\cdot10^{-5}$. The difference $F_{1,\mathrm{law}}-F_{1,\mathrm{base}}=-0.00032$ corresponds to $-0.032$ percentage points. Exact span extraction was slightly worse for the adapted model in this run. Lower token loss does not guarantee higher $F_1$ because these metrics evaluate different objects. One run per model cannot establish the stability of the difference or statistical equivalence.

In this split, 53\,660 of 53\,686 test spans (99.95\%) have the same normalized surface form and class in the training split. The denominator counts span occurrences, rather than unique forms: each occurrence of a recurring form is counted. This denominator describes coverage of the original annotations and is not substituted as the number of reference entities in the window-level $F_1$ evaluation. The remaining 26 spans are outside this coverage; no separate metric is reported for them. A matching form does not prove memorization, and an absent form does not guarantee semantic novelty.

Removing duplicate complete texts therefore did not produce a lexically disjoint span test. The high $F_1$ mainly concerns familiar forms under the window-level evaluation protocol. It does not measure the proportion of error-free documents or translate into a frequency of legal errors.

\begin{fixedfigure}
\centering
\begin{adjustbox}{max width=\linewidth}
\begin{tikzpicture}[font=\normalsize,box/.style={rounded corners,align=center,text width=5.5cm,inner sep=8pt,anchor=north}]
\node[box,draw=visTeal,fill=visTeal!8] (known) at (0,0.7) {\textcolor{visBlue}{53\,660 spans: familiar forms}\\form and class match\\the training split};
\node[box,draw=visAmber,fill=visAmber!9] (new) at (6.5,0.7) {\textcolor{visBlue}{26 spans: outside coverage}\\this form/class pair\\is absent from the training split};
\node[font=\large,text=visBlue] at (3.25,-1.5) {$53\,660+26=53\,686$ test spans};
\node[draw=visBlue,fill=visBlue!4,rounded corners,align=center,text width=11.5cm,inner sep=8pt] at (3.25,-2.9) {\textcolor{visBlue}{99.95\% of spans: familiar form and class.}\\High $F_1$ here barely tests transfer to new forms.\\Next test: separate training/test forms and repeat training.};
\end{tikzpicture}
\end{adjustbox}
\captionisp{Coverage of test spans by normalized form and class in training. Box areas are not proportional to counts. Familiar forms do not prove memorization, and an absent form does not imply entirely new meaning.}{Coverage of test spans by normalized form and class in training. Box areas are not proportional to counts. Familiar forms do not prove memorization, and an absent form does not imply entirely new meaning.}
\label{fig:span-coverage}
\end{fixedfigure}

\subhead{Evaluation boundaries}
The MLM result is restricted to the pair \base{}--\model{}, a fixed collection, and five masks. Semantic overlap between corpora, training variability, and transfer to other collections were not evaluated. Changing $L$ changes both available context and target tokens; the contribution of distant dependencies is not separated from prefix composition.

The NER test is almost entirely covered by familiar forms. Window-level \texttt{seqeval} evaluation without overlap merging is not equivalent to recovering character intervals in complete documents. There is no comparison with alternative encoders under an equal budget. Complexity estimates describe operation scaling rather than a measured advantage in runtime or total training memory.

\subhead{Reproducible evaluation protocol}
Reproducing the MLM comparison requires fixed model, tokenizer, and data versions, retained fragment identifiers, and inputs with identical masking and token replacements for each model pair. Losses are averaged first over hidden positions within a fragment, then equally over fragments and five masks according to Eqs.~\eqref{eq:external-estimator}--\eqref{eq:ci}. For NER, the evaluation unit is a window: exact token boundaries and entity class are checked in BIO sequences.

\section{Conclusion}
After continued pretraining of Russian ModernBERT on 304\,382 legislative documents, mean masked-token cross-entropy on the fixed external court-fragment collection was lower in all 15 combinations of three maximum input lengths and five masks. At $L=8192$, the mean paired difference was 0.06604 nats. The intervals describe sensitivity to masking on this collection; the causal contribution of distant context is not isolated.

In the window-level span extraction task, the adapted model had slightly lower $F_1$ than the baseline: $0.99820$ versus $0.99852$. Almost complete lexical coverage of the test split and a single run per model limit conclusions about transfer and the stability of this difference. The study provides evidence of improved MLM evaluation for the model pair examined, but establishes neither an NER advantage nor practical usefulness for legal expertise.

Figure~\ref{fig:evidence-map} summarizes the scope of the conclusions. Improved MLM evaluation, limited discriminative value of the NER test, and future research questions are kept separate; no direct causal transition between them is implied.

\begin{fixedfigure}
\centering
\begin{adjustbox}{max width=\linewidth}
\begin{tikzpicture}[node distance=6mm]
\node[prismdata,text width=0.78\linewidth] (reported)
  {\textcolor{visBlue}{Reported}\\MLM effect on a fixed court-fragment collection};
\node[prismresult,text width=0.78\linewidth,below=of reported] (limited)
  {\textcolor{visBlue}{Limited evidence}\\NER test mainly contains familiar forms; one fine-tuning run per model};
\node[prismbox,text width=0.78\linewidth,below=of limited] (unchecked)
  {\textcolor{visBlue}{Not tested}\\transfer to new forms; isolated distant-context contribution; practical legal utility};
\end{tikzpicture}
\end{adjustbox}
\captionisp{Evidence boundaries: reported language-modelling effect, limited entity-extraction evidence, and untested transfer, context and utility claims. The blocks do not imply a causal chain.}{Evidence boundaries: reported language-modelling effect, limited entity-extraction evidence, and untested transfer, context and utility claims. The blocks do not imply a causal chain.}
\label{fig:evidence-map}
\end{fixedfigure}

\section*{Acknowledgements}
The computations used equipment of the National Research Nuclear University MEPhI high-performance computing cluster.

\section*{Author contributions}
A. Kostin, S. Lagutin, F. Lashkin, T. Maksiyan, and D. Gvozdetsky contributed to constructing and reviewing the legal benchmark and advised on legal aspects of the study. V. Kirova, I. Kiselev, and R. Leva contributed to organizing the computing infrastructure, developing algorithmic solutions, and providing technical support for the experiments. I. Litvak provided scientific supervision, formulated the problem, and coordinated the study. I. Litvak and V. Volf contributed to designing the experiments, analyzing the results, and preparing the manuscript.

\section*{Funding}
The study received no external targeted funding.

\section*{Declaration of interests}
The authors declare no conflicts of interest.

\printbibliography[title={References}]

@misc{devlin2019bert,
 note = {Devlin J., Chang M.-W., Lee K., Toutanova K. BERT: Pre-training of Deep Bidirectional Transformers for Language Understanding~// Proc. NAACL-HLT. 2019. P.~4171--4186. DOI: 10.18653/v1/N19-1423.},
}

@misc{gururangan2020dapt,
 note = {Gururangan S., Marasovi\'c A., Swayamdipta S. et al. Don't Stop Pretraining: Adapt Language Models to Domains and Tasks~// Proc. ACL. 2020. P.~8342--8360. DOI: 10.18653/v1/2020.acl-main.740.},
}

@misc{chalkidis2020legalbert,
 note = {Chalkidis I., Fergadiotis M., Malakasiotis P., Aletras N., Androutsopoulos I. LEGAL-BERT: The Muppets straight out of Law School~// Findings of EMNLP. 2020. P.~2898--2904. DOI: 10.18653/v1/2020.findings-emnlp.261.},
}

@misc{niklaus2024multilegalpile,
 note = {Niklaus J., Matoshi V., Stürmer M., Chalkidis I., Ho D. MultiLegalPile: A 689GB Multilingual Legal Corpus~// Proc. ACL. 2024. P.~15077--15094. DOI: 10.18653/v1/2024.acl-long.805.},
}

@misc{saveliev2024corpus,
 note = {Saveliev D., Kuchakov R. The Russian Legislative Corpus. arXiv:2406.04855v3, 27.04.2026. DOI: 10.48550/arXiv.2406.04855.},
}

@misc{ruslawodcard,
 note = {Institute for the Rule of Law. RusLawOD: The Russian Legislative Corpus, 1991--2025. \url{https://huggingface.co/datasets/irlspbru/RusLawOD} (accessed: 30.07.2026).},
}

@misc{beltagy2020longformer,
 note = {Beltagy I., Peters M. E., Cohan A. Longformer: The Long-Document Transformer. 2020. arXiv:2004.05150. DOI: 10.48550/arXiv.2004.05150.},
}

@misc{xiao2021lawformer,
 note = {Xiao C., Hu X., Liu Z., Tu C., Sun M. Lawformer: A Pre-trained Language Model for Chinese Legal Long Documents~// AI Open. 2021. Vol.~2. P.~79--84. DOI: 10.1016/j.aiopen.2021.06.003.},
}

@misc{warner2024modernbert,
 note = {Warner B., Chaffin A., Clavi\'e B. et al. Smarter, Better, Faster, Longer: A Modern Bidirectional Encoder for Fast, Memory Efficient, and Long Context Finetuning and Inference. 2024. arXiv:2412.13663. DOI: 10.48550/arXiv.2412.13663.},
}

@misc{sudreshcard,
 note = {Lawful Good Project. Sud-resh benchmark. \url{https://huggingface.co/datasets/lawful-good-project/sud-resh-benchmark} (accessed: 30.07.2026).},
}

@misc{nercard,
 note = {TryDotAtwo. Russian Legal NER. \url{https://huggingface.co/datasets/TryDotAtwo/russian-legal-ner} (accessed: 05.10.2026).},
}

@misc{dao2023flashattention2,
 note = {Dao T. FlashAttention-2: Faster Attention with Better Parallelism and Work Partitioning. 2023. arXiv:2307.08691. DOI: 10.48550/arXiv.2307.08691.},
}

@misc{rajbhandari2020zero,
 note = {Rajbhandari S., Rasley J., Ruwase O., He Y. ZeRO: Memory Optimizations Toward Training Trillion Parameter Models~// Proc. SC20. 2020. P.~1--16. DOI: 10.1109/SC41405.2020.00024.},
}

@misc{loshchilov2019adamw,
 note = {Loshchilov I., Hutter F. Decoupled Weight Decay Regularization~// Proc. ICLR. 2019.},
}

@misc{repository,
 note = {TryDotAtwo. RuModernBERT-ruLaw: source code and experiment logs, branch master, commit b3b3cba1e03018a36e1c2e478f34fa3ca2bd0806. \url{https://github.com/TryDotAtwo/RuModernBERT-ruLaw} (source inspection: 01.10.2026).},
}

@misc{modelcard,
 note = {TryDotAtwo. RuModernBERT-ruLaw: model card. \url{https://huggingface.co/TryDotAtwo/RuModernBERT-ruLaw} (accessed: 05.10.2026).},
}
\end{document}